\batchmode
\documentclass[a4paper,12pt]{article}

\usepackage{comment}
\usepackage{fancyvrb}
\usepackage[round]{natbib}
\usepackage{graphicx}
\usepackage[ragged]{footmisc}
\usepackage{ragged2e}
\usepackage{hyperref}

\title{Simplification and integration in computing and cognition: the SP theory and the multiple alignment concept}

\author{J Gerard Wolff\footnote{Dr Gerry Wolff, BA (Cantab), PhD (Wales), CEng, MBCS (CITP); CognitionResearch.org, 18 Penlon, Menai Bridge, Anglesey, LL59 5LR, UK; \href{mailto:jgw@cognitionresearch.org}{jgw@cognitionresearch.org}; +44 (0) 1248 712962; +44 (0) 7746 290775;  \href{http://www.cognitionresearch.org}{www.cognitionresearch.org}.}}

\normalsize

\date{30 November 2012}

\begin{document}

\bibliographystyle{plainnat}

\maketitle

\begin{abstract}

The main purpose of this article is to describe potential benefits and applications of the SP theory, a unique attempt to simplify and integrate ideas across artificial intelligence, mainstream computing and human cognition, with information compression as a unifying theme.
The theory, including a concept of multiple alignment, combines conceptual simplicity with descriptive and explanatory power in several areas including representation of knowledge, natural language processing, pattern recognition, several kinds of reasoning, the storage and retrieval of information, planning and problem solving, unsupervised learning, information compression, and human perception and cognition.

In the {\em SP machine}---an expression of the SP theory which is currently realised in the form of computer models---there is potential for an overall simplification of computing systems, including software.

As a theory with a broad base of support, the SP theory promises useful insights in many areas and the integration of structures and functions, both within a given area and amongst different areas. There are potential benefits in natural language processing (with potential for the understanding and translation of natural languages), the need for a versatile intelligence in autonomous robots, computer vision, intelligent databases, maintaining multiple versions of documents or web pages, software engineering, criminal investigations, the management of big data and gaining benefits from it, the semantic web, medical diagnosis, the detection of computer viruses, the economical transmission of data, and data fusion.
Further development of these ideas would be facilitated by the creation of a high-parallel, web-based, open-source version of the SP machine, with a good user interface. This would provide a means for researchers to explore what can be done with the system and to refine it.

\end{abstract}


\section{Introduction}

Since about 1987, I have been fascinated by the idea that it might be possible to create a theory that simplifies and integrates ideas across artificial intelligence, mainstream computing and human cognition, with information compression as a unifying theme. The first rather crude attempt at such a theory, with the main focus on computing, was described in \citet{wolff_1990}. In the following years, progressively more refined versions of the `SP' theory have been described in several peer-reviewed journal articles and, in some detail, in a book \citep{wolff_2006}.\footnote{Bibliographic details of relevant publications may be found via links from \url{www.cognitionresearch.org/sp.htm}.}

The main purpose of this article is to describe potential benefits and applications of the theory. In brief:

\begin{itemize}

\item The SP theory combines conceptual simplicity with descriptive and explanatory power.

\item It has potential to simplify computing systems, including software.

\item As a theory with a broad base of support, the SP theory promises to provide useful insights in many areas and is likely to facilitate the simplification and integration of structures and functions, both within a given area and amongst different areas.

\end{itemize}

These points will be discussed, each in its own section, below. But first, as a background and context to the discussion, the next section considers a little philosophy and outlines the SP theory, the multiple alignment concept, and associated ideas.

\section{Background and context}

\subsection{A little philosophy}\label{philosophy_section}

Amongst alternative criteria for the success of a theory, what appears to be the most widely accepted is the principle (`Occam's Razor') that a good theory should combine simplicity with descriptive or explanatory power.\footnote{According to this criterion, a good theory in a given area lies somewhere between things which are simple but very general---such as the laws of arithmetic---and over-specific `theories' that simply repeat what it is they are meant to explain.} This equates with the idea---which echoes the underlying theme of the SP theory itself\footnote{Hence the name `SP'.}---that a good theory should compress empirical data via a reduction of `redundancy' in the data (thus increasing its `simplicity'), whilst retaining as much as possible of its non-redundant descriptive `power'. As John Barrow has written: ``Science is, at root, just the search for compression in the world.'' \citep[p. 247]{barrow_1992}.

These principles are prominent in most areas of science: the Copernican heliocentric theory of the Sun and the planets is rightly seen to provide a simpler and more powerful account of the facts than Ptolemy's Earth-centred theory, with its complex epicycles; physicists have a keen interest in the possibility that quantum theory and relativity might be unified; biology would be greatly impoverished without modern understandings of evolution, genetics, and DNA; and so on.

But in research in computer science, including artificial intelligence, there is a fragmentation of the field into a myriad of concepts and many specialisms, with little in the way of an overarching theory to pull everything together.\footnote{Similar things were said by Allen Newell about research in human perception and cognition, in his well-known essay on why ``You can't play 20 questions with nature and win'' (\citet{newell_1973}. See also \citet{newell_1992,newell_1990}).} And some areas, such as computer vision, appear to be overburdened with over-complex mathematics.\footnote{Of course, mathematics is often very useful---the SP theory has a solid mathematical foundation---but large amounts of mathematics suggest that the ideas are on the wrong track, somewhat like Ptolemy's epicycles. And the limitations of mathematical notations can be an obstacle to the development of other concepts that do not fit easily into a mathematical perspective.}

Perhaps the Turing machine model of computing \citep{turing_1936} provides what is needed? That theory has of course been brilliantly successful, providing theoretical underpinnings for the many useful things that can be done with computers. But it does not solve the problem of fragmentation and, although Alan Turing recognised that computers might become intelligent \citep{turing_1950}, the Turing theory, in itself, does not tell us how!

Whether or not the SP theory succeeds in plugging these gaps, I believe there is a pressing need, in computer science and AI, for a stronger focus on the simplification and integration of ideas.\footnote{Notice that collaboration amongst different specialisms---which is necessary or useful for some kinds of projects---does not in itself achieve simplification and integration of ideas.} In all areas, including quite specialised areas, a theory that scores well in terms of simplicity and power is, compared with any weaker theory, likely to yield deeper insights and better integration of concepts, both within a given area and amongst different areas (see also Section \ref{areas_of_application}).

\subsection{Outline of the SP theory}\label{sp_outline}

\subsubsection{Origins and motivation}\label{origins_motivation}

Much of the inspiration for the SP theory is a body of research, pioneered by Fred Attneave \citep[see, for example,][]{attneave_1954}, Horace Barlow \citep[see, for example,][]{barlow_1959,barlow_1969}, and others, showing that many aspects of the workings of brains and nervous systems may be understood as compression of information. For example, when we view a scene with two eyes, the image on the retina of the left eye is almost exactly the same as the image on the retina of right eye, but our brains merge the two images into a single percept, and thus compress the information \citep{barlow_1969}.\footnote{This focus on compression of information in binocular vision is distinct from the more usual interest in the way that slight differences between the two images enables us to see the scene in depth.}

The theory also draws on principles of `minimum length encoding' pioneered by \citet{solomonoff_1964,wallace_boulton_1968,rissanen_1978}, and others.
The close connection that exists between information compression and concepts of probability \citep[see, for example,][]{li_vitanyi_2009} is an important motivation for the theory.

These ideas, with several observations about computing and AI, suggested that it might be possible to develop a theory bridging artificial intelligence, mainstream computing and human cognition, with information compression as a unifying theme \citep[Chapter 2]{wolff_2006}.

\subsubsection{Elements of the theory}

The main elements of the SP theory are:

\begin{itemize}

\item All knowledge is expressed as arrays of atomic symbols ({\em patterns}) in one or two dimensions.

\item Patterns that represent data coming in from the environment are classified as `New'. Patterns representing stored knowledge are classified as `Old'.

\item The system is designed for the unsupervised learning of Old patterns by compression of New patterns.

\item An important part of this process is the economical encoding of New patterns in terms of Old patterns, if they exist. This may be seen to achieve such things as pattern recognition, parsing or understanding of natural language, or other kinds of interpretation of incoming information in terms of stored knowledge, including several kinds of reasoning.

\item Compression of information is achieved via the matching and unification (merging) of patterns, with an improved version of dynamic programming \citep[Appendix A]{wolff_2006} providing flexibility in matching, and with key roles for the frequency of occurrence of patterns, and their sizes.\footnote{In the SP theory, the matching and unification of patterns is seen as being closer to the bedrock of information compression, and with greater heuristic value, than more mathematical techniques such as wavelets, arithmetic coding, or the like. The mathematics which provides the basis for those techniques may itself be founded on the matching and unification of patterns \citep[see][Chapter 10]{wolff_2006}.}

\item The concept of {\em multiple alignment}, outlined in Section \ref{multiple_alignment}, is a powerful central idea, similar to the concept of multiple alignment in bioinformatics but with important differences.

\item Owing to the intimate connection, already mentioned, between information compression and concepts of probability, the SP theory provides for the calculation of probabilities for inferences made by the system, and probabilities for parsings, recognition of patterns, and so on.

\end{itemize}

\subsubsection{The SP62 and SP70 computer models}

The SP theory is currently expressed in two computer models, SP62 and SP70. Although SP62 is a subset of SP70 which can build multiple alignments but which lacks any capability for unsupervised learning, it has been convenient to maintain them as separate models.

The SP models and their precursors have played a key part in the development of the theory:

\begin{itemize}

\item By forcing precision in the definition of the theory where there might otherwise be vagueness.

\item By providing a convenient means of encoding the simple but important mathematics that underpins the SP theory, and performing relevant calculations, including calculations of probability.

\item By providing a means of seeing quickly the strengths and weaknesses of proposed mechanisms or processes.

\item By providing a means of demonstrating what can be achieved with the theory.

\end{itemize}

\subsubsection{The SP machine}\label{sp_machine}

The SP models may be regarded as implementations of the {\em SP machine}, an expression of the SP theory and a means for it to be applied.

A useful step forward in the development of the SP theory and the SP machine would be the creation, as an open-source software virtual machine, of a high-parallel, web-based version, taking advantage of powerful search mechanisms in one of the existing search engines, and with a good user interface. This would provide a means for researchers to explore what can be done with the system and to refine it.

Further ahead, there may be a case for the creation of new hardware, dedicated to the building of multiple alignments and other processes in the SP framework.

\subsubsection{The multiple alignment concept}\label{multiple_alignment}

A concept of multiple alignment is widely used in bioinformatics where it means a process of arranging, in rows or columns, two or more DNA sequences or amino-acid sequences so that matching symbols---as many as possible---are aligned orthogonally in columns or rows.

\begin{figure}[!hbt]
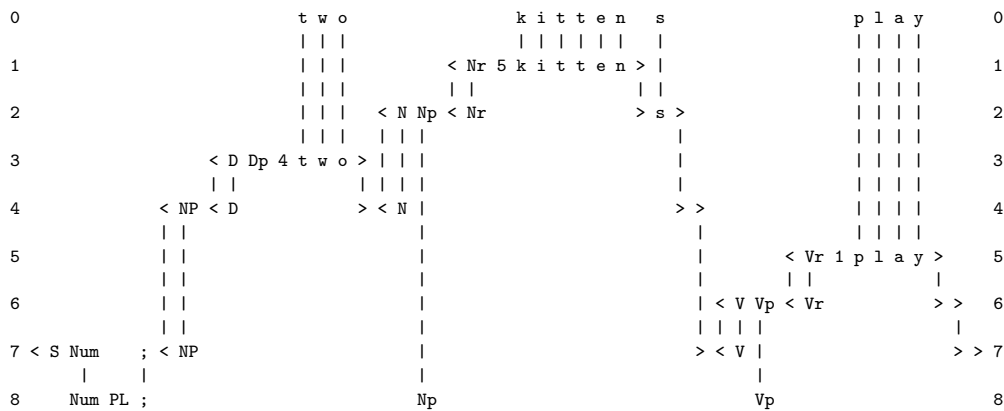

\fontsize{07.50pt}{09.00pt}
\centering
{\bf
\begin{BVerbatim}
0                            t w o                 k i t t e n   s                   p l a y       0
                             | | |                 | | | | | |   |                   | | | |
1                            | | |          < Nr 5 k i t t e n > |                   | | | |       1
                             | | |          | |                | |                   | | | |
2                            | | |   < N Np < Nr               > s >                 | | | |       2
                             | | |   | | |                         |                 | | | |
3                   < D Dp 4 t w o > | | |                         |                 | | | |       3
                    | |            | | | |                         |                 | | | |
4              < NP < D            > < N |                         > >               | | | |       4
               | |                       |                           |               | | | |
5              | |                       |                           |        < Vr 1 p l a y >     5
               | |                       |                           |        | |            |
6              | |                       |                           | < V Vp < Vr           > >   6
               | |                       |                           | | | |                   |
7 < S Num    ; < NP                      |                           > < V |                   > > 7
       |     |                           |                                 |
8     Num PL ;                           Np                                Vp                      8
\end{BVerbatim}
}
\caption{The best multiple alignment created by the SP62 model with a store of Old patterns like those in rows 1 to 8 (representing grammatical structures, including words) and the New pattern (representing a sentence to be parsed) shown in row 0.}
\label{two_kittens_play}
\end{figure}

\begin{figure}[!hbt]
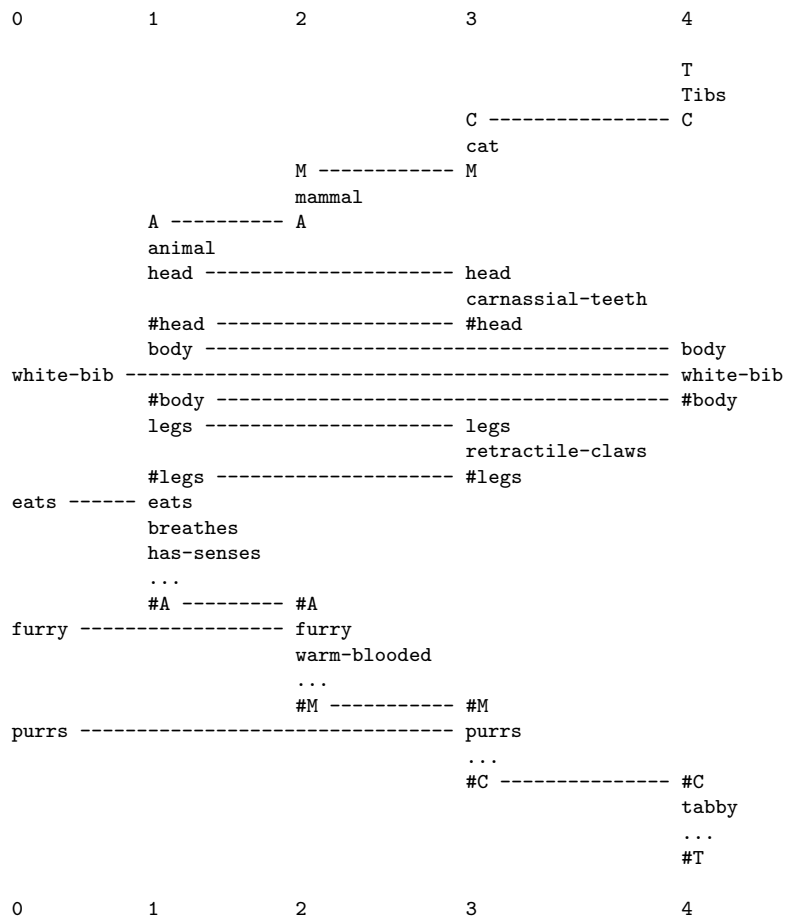

\fontsize{08.00pt}{09.60pt}
\centering
{\bf
\begin{BVerbatim}
0           1            2              3                  4

                                                           T
                                                           Tibs
                                        C ---------------- C
                                        cat
                         M ------------ M
                         mammal
            A ---------- A
            animal
            head ---------------------- head
                                        carnassial-teeth
            #head --------------------- #head
            body ----------------------------------------- body
white-bib ------------------------------------------------ white-bib
            #body ---------------------------------------- #body
            legs ---------------------- legs
                                        retractile-claws
            #legs --------------------- #legs
eats ------ eats
            breathes
            has-senses
            ...
            #A --------- #A
furry ------------------ furry
                         warm-blooded
                         ...
                         #M ----------- #M
purrs --------------------------------- purrs
                                        ...
                                        #C --------------- #C
                                                           tabby
                                                           ...
                                                           #T

0           1            2              3                  4
\end{BVerbatim}
}
\caption{A multiple alignment showing how an unknown creature, with features shown in the New pattern in column 0, may be recognised in terms of Old patterns at multiple levels of abstraction: as an animal (column 1), a mammal (column 2), a cat (column 3), and as a specific individual, `Tibs' (column 4).}
\label{recognise_cat}
\end{figure}

In the SP theory, this concept has been borrowed and adapted, as illustrated in Figures \ref{two_kittens_play} and \ref{recognise_cat}. The main difference between the two concepts is that, in bioinformatics, all sequences have the same status, whereas in the SP theory, the system attempts to create a multiple alignment which enables one New pattern\footnote{In some applications, there may be more than one New pattern in a multiple alignment.} (shown in row 0 in Figure \ref{two_kittens_play} and column 0 in Figure \ref{recognise_cat}) to be encoded economically in terms of one or more Old patterns (shown in rows 1 to 8 in Figure \ref{two_kittens_play} and in columns 1 to 4 in Figure \ref{recognise_cat}).

In Figure \ref{two_kittens_play}, the New pattern is a sentence, the Old patterns represent grammatical structures (including words), and the whole multiple alignment may be seen as a parsing of the sentence in terms of a grammar.

Notice how, in row 8, the system marks the dependency between the plural subject of the sentence (`Np') and the plural main verb (`Vp').

Although this is not illustrated here, the system also supports the production of language as well as its analysis, it facilitates the integration of syntax and semantics, and it is robust in the face of errors in the sentence to be parsed or the stored grammatical patterns, or both.

As an indication of the versatility of the multiple alignment concept, Figure \ref{recognise_cat} illustrates how it may model recognition of an unknown creature at multiple levels of abstraction, as described in the caption to the figure.

Not illustrated here is the way the framework can accommodate cross-classification and the integration of class hierarchies with part-whole hierarchies, and how recognition may be achieved despite errors in incoming data or stored knowledge, or both.

These are just two of several things that can be modelled within the multiple alignment framework. Others are summarised in Section \ref{sp_simplicity_power}, below.

\subsubsection{A possible neural basis for the SP theory}\label{neural_basis}

\begin{figure}[!hbt]
\centering
\includegraphics[width=0.6\textwidth]{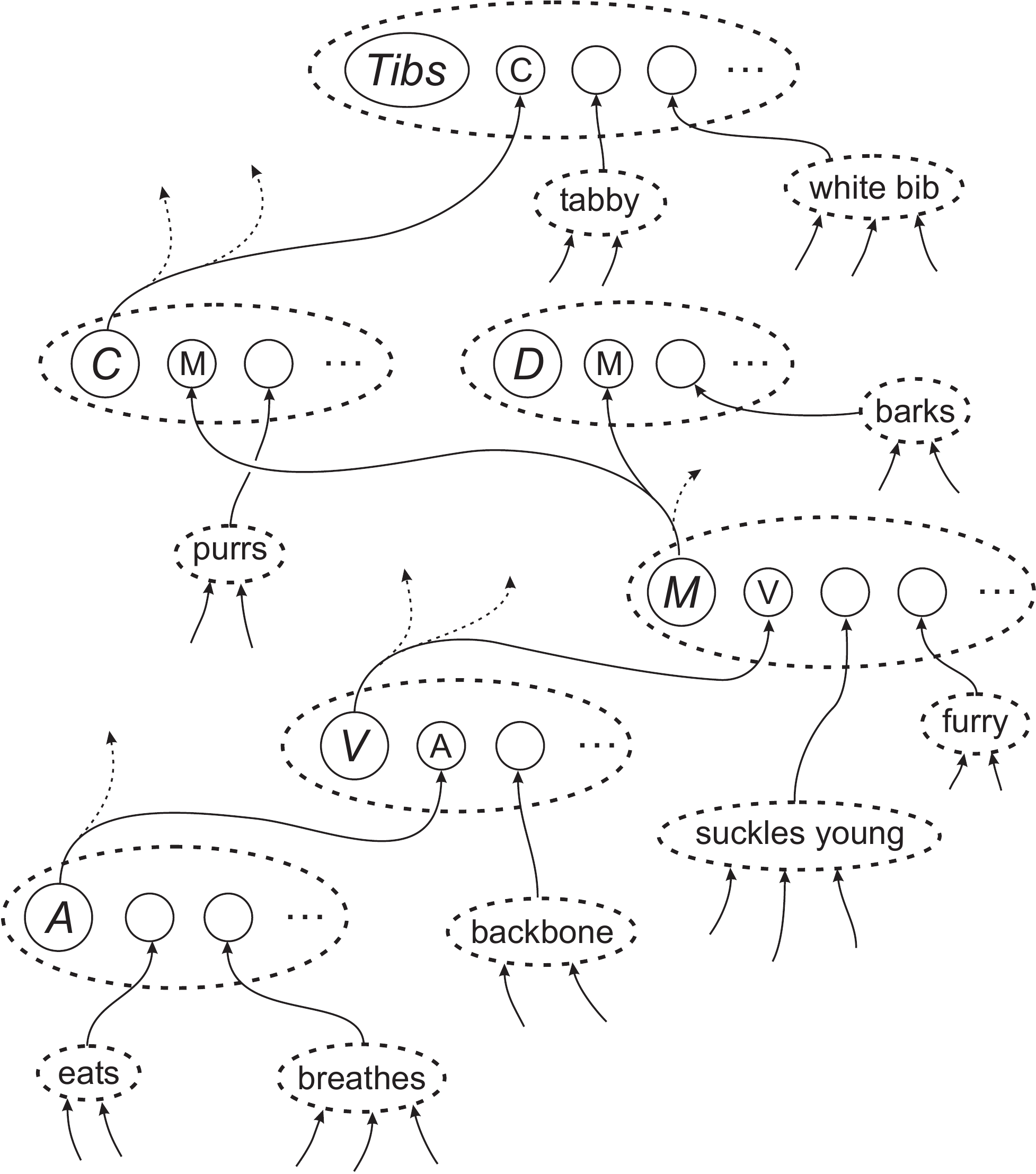}
\caption{Schematic representation of hypothesised neural analogues of SP patterns and their inter-connections. Reproduced from Figure 11.6 of \protect\citet{wolff_2006}, with permission.}
\label{class_part_figure}
\end{figure}

A tentative part of the SP theory is the idea that the cortex of the brains of mammals---which is, topologically, a two-dimensional sheet---may be, in some respects, like a sheet of paper on which neural analogues of SP patterns may be written---as shown schematically in Figure \ref{class_part_figure}. Unlike information written on a sheet of paper, there are neural connections between patterns---as shown in the figure---and communications amongst them.

\subsubsection{Related research}\label{related_research}

To my knowledge, the SP theory is unique in attempting to simplify and integrate ideas across artificial intelligence, mainstream computing, and human cognition.

But there has for some time been an interest in unified theories of cognition such as SOAR \citep[see, for example,][]{laird_2012} and ACT-R \citep[see, for example,][]{anderson_etal_2004}.\footnote{Both of them inspired by Allen Newell \citep[eg,][]{newell_1973}.} And, although it is still very much a minority interest amongst researchers in artificial intelligence, there has in recent years been renewed interest in artificial intelligence as a whole \citep[see, for example,][]{agi_2011}.

As mentioned in Section \ref{origins_motivation}, the SP theory is inspired in part by research showing the importance of information compression in brains and nervous systems, and it draws on other research on principles of minimum length encoding, and the close relation that exists between information compression and concepts of probability.

\subsubsection{Unfinished business}\label{unfinished_business}

Like most theories, the SP theory cannot claim to solve every problem in its target areas. The main shortcomings of the theory as it stands now are:

\begin{itemize}

\item {\em Processing of information in two or more dimensions}. No attempt has yet been made to generalise the SP models to work with patterns in two dimensions, although that appears to be feasible to do, as outlined in \citet[Section 13.2.1]{wolff_2006}. As noted in \citet[Section 13.2.2]{wolff_2006}, it is possible that information with dimensions higher than two may be encoded in terms of patterns in one or two dimensions, somewhat in the manner of architects' drawings. With the aid of triangulation to determine depths, a 3D structure may be stitched together from several partially-overlapping 2D views, in much the same way that, in digital photography, a panoramic view may be created from partially-overlapping pictures \citep[For reviews of related work in computer vision, see][]{prince_2012,szeliski_2011}.

\item {\em Recognition of perceptual features in speech and visual images}. For the SP system to be effective in the processing of speech or visual images, it seems likely that some kind of preliminary processing will be required to identify low level perceptual features such as, in the case of speech, phonemes, formant ratios, or formant transitions, or, in the case of visual images, edges, angles, colours, luminances, or textures. In vision, at least, it seems likely that the SP framework itself will prove relevant since edges may be seen as zones of non-redundant information between uniform areas containing more redundancy and, likewise, angles may be seen to provide significant information where straight edges, with more redundancy, come together \citep[See, for example,][]{attneave_1954}. As a stop-gap solution, the preliminary processing may be done using existing techniques for the identification of low-level perceptual features \citep[see, for example,][Chapter 13]{prince_2012}.

\item {\em Unsupervised learning}. A limitation of the SP70 model as it is now is that it cannot learn intermediate levels of abstraction in grammars or the like, and it cannot learn discontinuous dependencies in natural language syntax or other kinds of information. I believe these problems are soluble and that solving them will greatly enhance the capabilities of the system for the unsupervised learning of structure in data.

\item {\em Processing of numbers}. The SP models work with atomic symbols such as ASCII characters or strings of characters with no intrinsic meaning. In itself, the SP system does not recognise the arithmetic meaning of numbers such as `37' or `652' and will not process them correctly. However, the system has the potential to handle mathematical concepts correctly, if it is supplied with the relevant mathematical rules \citep[Chapter 10]{wolff_2006}. As a stop-gap solution, existing technologies may provide whatever arithmetic processing may be required.

\end{itemize}

\section{The SP theory combines simplicity with descriptive and explanatory power}\label{sp_simplicity_power}

In principle, it should be possible to evaluate scientific theories quite precisely in terms of simplicity and power, and to compare one theory with another, by measuring their ability to compress empirical data. But techniques and technologies are not yet well enough developed to make this feasible, and there is in any case the difficulty that it is rare for rival theories to address precisely the same body of empirical data. So in evaluating theories we still need to rely on more-or-less informal judgements of simplicity and explanatory or descriptive power.

In those terms, the SP theory appears to score well:

\begin{itemize}

\item The SP theory, including the multiple alignment concept, is not trivially simple but it is not unduly complicated either. The SP70 model, which is the most comprehensive expression of the theory as it stands now, is embodied in an `exec' file requiring less than 500 KB of storage space.

\item Largely because of the versatility of the multiple alignment concept, the SP theory can model a good range of concepts and phenomena in artificial intelligence, mainstream computing, and human cognition, as summarised in the following subsections.

\end{itemize}

\subsection{The SP theory as a theory of computing}

In \citet[Chapter 4]{wolff_2006}, I have argued that the operation of a Post canonical system \citep{post_1943} may be interpreted in terms of the SP theory and that, since it is accepted that the concept of a Post canonical system is equivalent to the Turing machine model of computing (which is itself widely regarded as a definition of the concept of `computing'), the SP theory may itself be regarded as a theory of computing---and not merely a theory of human cognition or artificial intelligence. By contrast with the Turing model and the Post canonical system, the fundamentals of the SP theory are multiple alignment and the compression of information by the matching and unification of patterns.

It is pertinent to mention here that the SP theory also has what I believe to be some useful things to say about the nature of mathematics \citep[see][Chapter 10]{wolff_2006}.

\subsection{Representation of knowledge}\label{representation_of_knowledge}

Despite the simplicity of representing knowledge with patterns, the way they are processed within the multiple alignment framework gives them the versatility to represent several kinds of knowledge, including grammars for natural languages, class hierarchies, part-whole hierarchies, discrimination networks, relational tuples, if-then rules, and others mentioned elsewhere in this article. This versatility is achieved without recourse to the kinds of constructs used in, for example, the `OWL 2' web ontology language:\footnote{See, for example, the {\em OWL 2 Web Ontology Language Primer} at \url{www.w3.org/TR/owl2-primer}.} `SubClassOf', `ObjectProperty', `SubObjectPropertyOf', and so on.

One universal format for knowledge and one universal framework for processing means that any subset of the different kinds of knowledge may be combined flexibly and seamlessly according to need.

\subsection{Natural language processing}\label{nl_processing}

One of the main strengths of the SP system is in natural language processing:

\begin{itemize}

\item As illustrated in Figure \ref{two_kittens_play}, grammatical rules, including words and their grammatical markers, may be represented with SP patterns.

\item Both the parsing and production of natural language may be modelled via the building of multiple alignments.

\item The framework provides an elegant means of representing discontinuous dependencies in syntax, including overlapping dependencies such as number dependencies and gender dependencies in languages like French.\footnote{As an illustration, row 8 of Figure \ref{two_kittens_play} enforces the rule that a plural subject for a sentence (`Np') must be followed by a plural main verb (`Vp').}

\item The system may also model non-syntactic `semantic' structures such as class-inclusion hierarchies and part-whole hierarchies (as noted in Section \ref{representation_of_knowledge} and elsewhere).

\item The system supports the seamless integration of syntax with semantics, providing a role for non-syntactic knowledge in the processing of natural language, and giving potential for the understanding of natural languages and interlingua-based translations amongst languages.

\item The system is robust in the face of errors of omission, commission or substitution in sentences to be analysed, or stored linguistic knowledge, or both.

\item The importance of context in the processing of language \citep[see, for example,][]{iwanska_zadrozny_1997} is accommodated in the way the system searches for a global best match for patterns: any pattern or partial pattern may be a context for any other.

\end{itemize}

\subsection{Pattern recognition}\label{pattern_recognition}

The system provides a powerful framework for pattern recognition:

\begin{itemize}

\item It can model pattern recognition at multiple levels of abstraction, as outlined in Section \ref{multiple_alignment}.

\item As noted in Section \ref{representation_of_knowledge}, the system also provides for cross-classification and the integration of class-inclusion relations with part-whole hierarchies.

\item The system provides for seamless integration of pattern recognition with various kinds of reasoning (Section \ref{reasoning_section}). As a simple example, if Tibs is recognised as a mammal on the strength of his or her furry coat, we may infer from the relevant pattern that the creature is warm blooded.

\item A probability may be calculated for any given classification or any associated inference.

\item The system is robust in the face of errors of omission, commission or substitution in incoming data, or stored knowledge, or both.

\item As in the processing of natural language (Section \ref{nl_processing}), the importance of context in recognition \citep[see, for example,][]{oliva_torralba_2007} is accommodated in the way the system searches for a global best match for patterns.

\end{itemize}

\subsection{Reasoning}\label{reasoning_section}

The SP system can model several kinds of reasoning including one-step `deductive' reasoning, abductive reasoning, reasoning with probabilistic decision networks and decision trees, reasoning with `rules', nonmonotonic reasoning and reasoning with default values, reasoning in Bayesian networks (including `explaining away'), causal diagnosis, and reasoning which is not supported by evidence.

Since these several kinds of reasoning all flow from one computational framework (multiple alignment), they may be seen as aspects of one process, working individually or together without awkward boundaries.

Plausible lines of reasoning may be achieved, even when relevant information is incomplete.

Probabilities of inferences may be calculated, including extreme values (0 or 1) in the case of logic-like `deductions'.

\subsection{Information storage and retrieval}\label{info_storage_retrieval}

The SP theory provides a versatile model for database systems, with the ability to accommodate object-oriented structures (with class hierarchies and inheritance of attributes, including multiple inheritance), as well as relational `tuples', and network and tree models of data \citep{wolff_sp_intelligent_database}. Unlike most database systems, it provides a simple and elegant means of combining class hierarchies with part-whole hierarchies.

It lends itself most directly to information retrieval in the manner of query-by-example but it appears to have potential to support the use of query languages such as SQL.

\subsection{Planning and problem solving}\label{planning_problem_solving}

The SP framework provides a means of planning a route between two places, and, with the translation of geometric patterns into textual form, it can solve the kind of geometric analogy problem that may be seen in some puzzle books and IQ tests \citep[Chapter 8]{wolff_2006}.

\subsection{Unsupervised learning}\label{unsupervised learning}

The SP70 model can derive a plausible grammar from a set of sentences without supervision or error correction by a `teacher', without the provision of `negative' samples, and without the grading of samples from simple to complex. It thus overcomes restrictions on what can be achieved with some other models of learning \citep[eg,][]{gold_1967}, and reflects more accurately what is known about how children learn their first language or languages \citep{wolff_1988}.

\begin{figure}[!hbt]
\centering
\includegraphics[width=85mm,height=114mm]{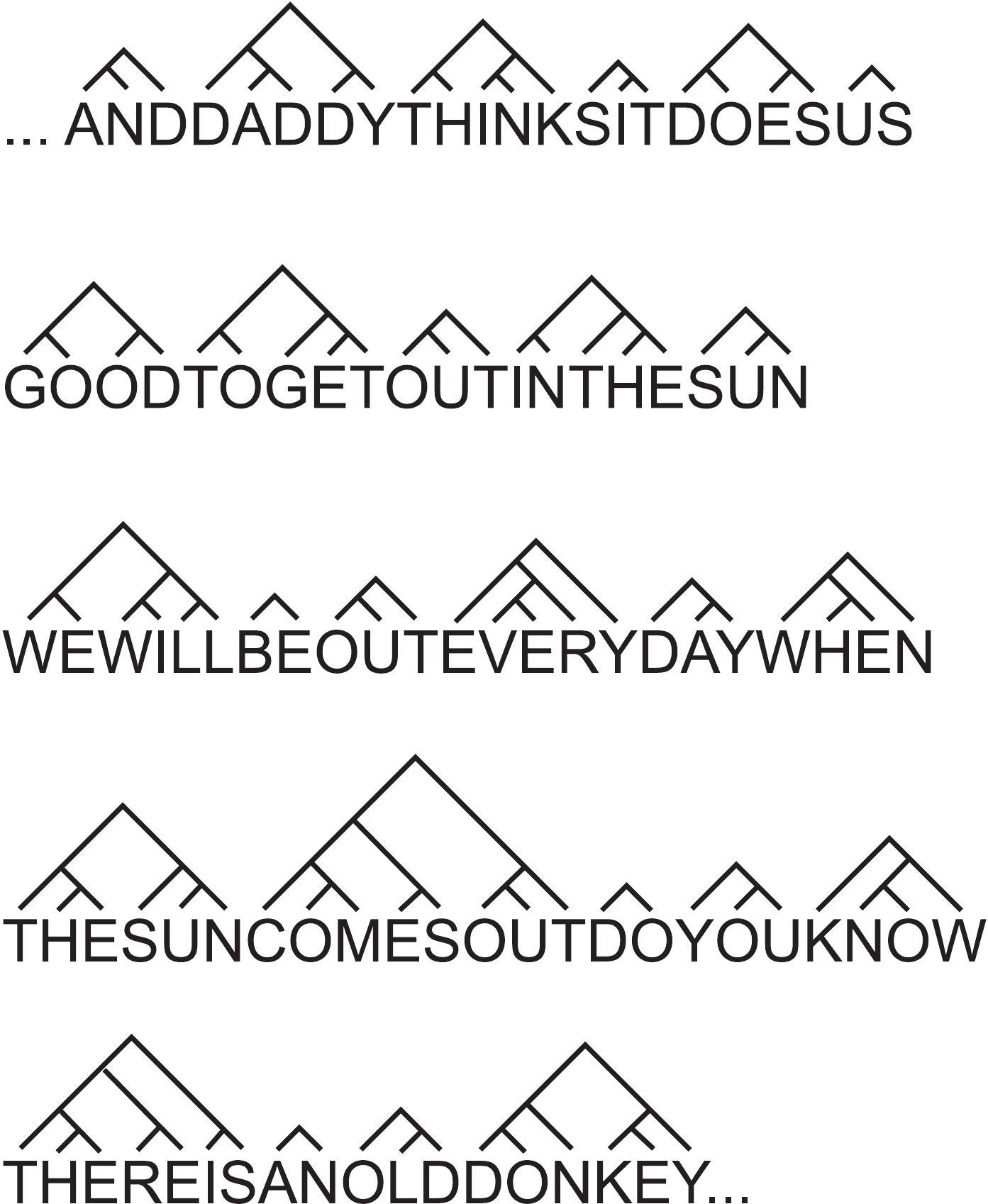}
\caption{Part of a parsing created by program MK10 \citep{wolff_1977} from a 10,000 letter sample of English (book 8A of the Ladybird Reading Series) with all spaces and punctuation removed. The program derived this parsing from the sample alone, without any prior dictionary or other knowledge of the structure of English. Reproduced from Figure 7.3 of \protect\citet{wolff_1988}, with permission.}
\label{discovery_of_words}
\end{figure}

The model draws on earlier research showing that inductive learning via principles of `minimum length encoding' (the matching and unification of patterns) can lead to the discovery of entities---such as words in natural languages or `objects' in the world of vision---that are psychologically natural, as illustrated in Figure \ref{discovery_of_words}. This principle---the discovery of natural structures via information compression (DONSVIC)---is also likely to apply to the discovery of classes of entity.

As was mentioned in Section \ref{unfinished_business}, the model, as it stands now, is not able to derive intermediate levels of abstraction or discontinuous dependencies in data, but those problems appear to be soluble.

\subsection{Compression of information}\label{compression_potential}

Since information compression is central to the workings of the SP system, it is natural to consider whether the system might provide useful insights in that area. The potential of the system to detect and encode discontinuous dependencies in data suggests that it may be able to extract kinds of redundancy in information that are not accessible via standard methods for the compression of information.

In terms of the trade-off that exists between computational resources that are required and the level of compression that can be achieved, it seems likely that the system will be towards the `up market' end of the spectrum---by contrast with LZW algorithms and the like, which have been designed to be `quick-and-dirty', sacrificing performance for speed on low-powered computers.

\subsection{Human perception and cognition}

Since much of the inspiration for the SP theory has come from evidence, mentioned in Section \ref{sp_outline}, that, to a large extent, the workings of brains and nervous systems may be understood in terms of information compression, the theory is about perception and cognition as well as artificial intelligence and mainstream computing.

That said, the main elements of the theory---the multiple alignment concept in particular---are theoretical constructs derived from what appears to be necessary to model, in an economical way, such things as pattern recognition, reasoning, and so on. In \citet[Chapter 11]{wolff_2006}, I have described in outline how such things as SP patterns and multiple alignments may be realised with neurons and connections between them. Given the enormous complexity of the brains of humans and other mammals, and given the difficulties of working with living brains, it is likely to be difficult to obtain direct evidence for or against the kinds of neural structures that I have described. But, because of the strength of support for the constructs in the SP theory, I believe that area of research deserves attention.

\section{How the SP theory may promote simplification of computing systems}\label{simplification_section}

Apart from the potential of the SP theory to simplify concepts in artificial intelligence and mainstream computing (Section \ref{sp_simplicity_power}), it can help to simplify computing systems, including software.

The principle to be described here is already familiar in the way databases and expert systems are structured.

Early databases were each written as a monolithic system containing everything needed for its operation, and likewise for expert systems. But it soon became apparent that, in the case of databases, a lot of effort could be saved by creating a generalised `database management system' (DBMS), with a user interface and mechanisms for the storage and retrieval of information, and creating new databases by loading the DBMS with different bodies of information, according to need. In a similar way, an `expert system shell', with a user interface and with mechanisms for inference and for the storage and retrieval of information, eliminates the need to create those facilities repeatedly in each new expert system.

The SP system takes this principle further. It aims to provide a general-purpose `intelligence'---chiefly the multiple alignment framework---and thus save the need to create those kinds of mechanisms repeatedly in different AI applications: pattern recognition, natural language processing, several kinds of reasoning, planning, problem solving, unsupervised learning, and more.

The principle extends to conventional applications, since the matching and unification of patterns, and a process of searching amongst alternative matches for patterns (which are elements of the multiple alignment framework), are used in many kinds of application outside the world of artificial intelligence.

\begin{figure}[!hbt]
\centering
\includegraphics[width=0.9\textwidth]{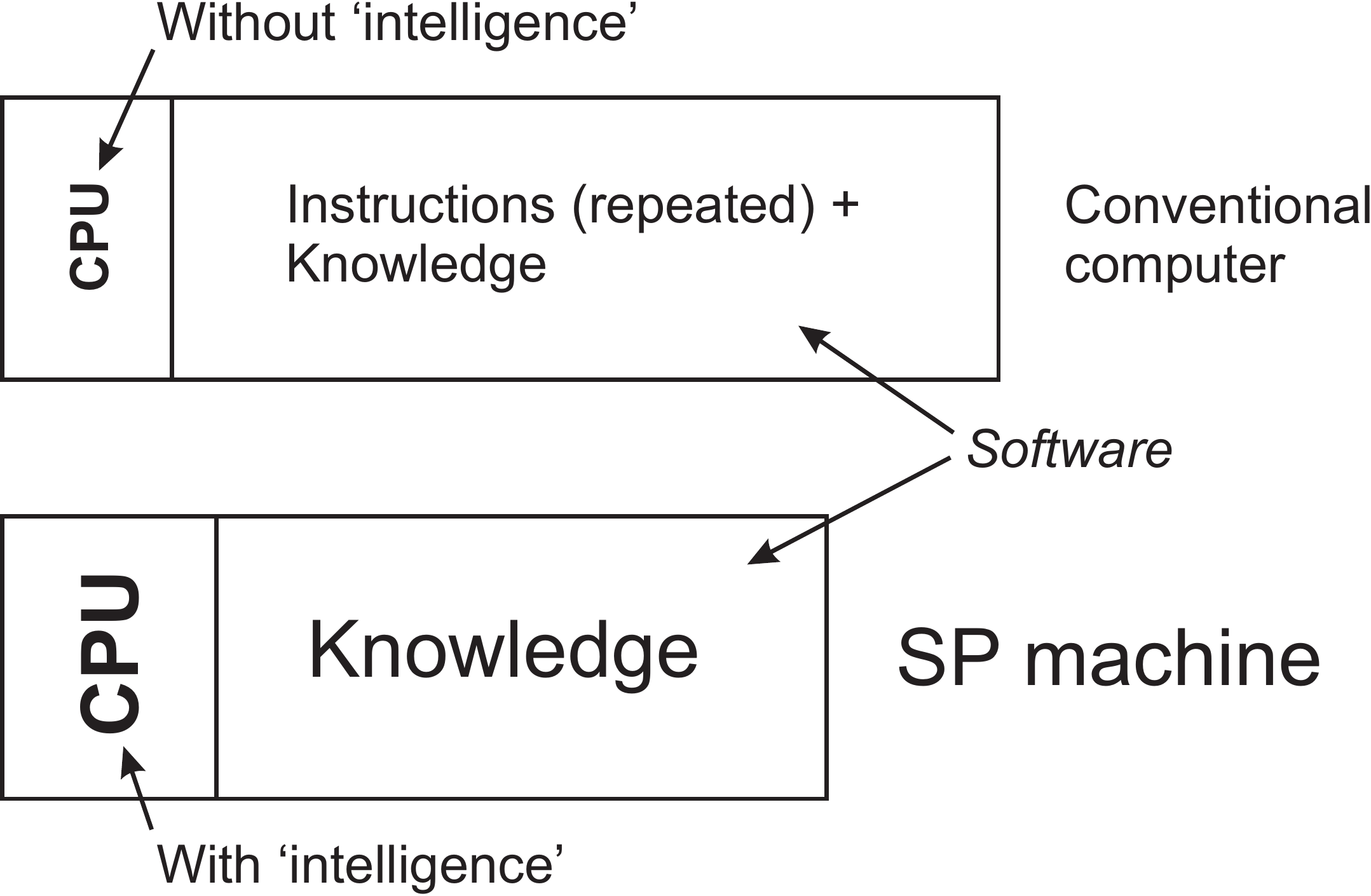}
\caption{Schematic representations of a conventional computer and an SP machine, as discussed in the text.}
\label{computers_schematic}
\end{figure}

The way in which this principle may help to simplify computing systems is shown schematically in Figure \ref{computers_schematic}.

In a conventional computer, shown at the top of the figure, there is a central processing unit (`CPU') which is relatively simple and without `intelligence'.

Software in the system, shown to the right, is a combination of two things:

\begin{itemize}

\item Domain-specific knowledge such as knowledge of accountancy, geography, the organisation and procedures of a business, and so on.\footnote{Even in systems such as DBMSs or expert-system shells, mentioned above, the domain-specific knowledge which is loaded into the system may be regarded as part of the software.}

\item Processing instructions to provide the intelligence that is missing in the CPU, chiefly processes for the matching and unification of patterns, for searching amongst alternative matches to find one or more that are `good', and the implicit formation of multiple alignments. These kinds of processing instructions---let us call them `MUP' instructions---are required in many different kinds of application, meaning that there is a considerable amount of redundancy in conventional computing systems---where the term `system' includes both the variety of programs that may be run, as well as the hardware on which they run.

\end{itemize}

In the SP machine, shown schematically at the bottom of the figure, the CPU is more complicated (as suggested by the larger size of `CPU' in the figure) and provides the `intelligence' that is missing in the CPU of a conventional computer. This should mean that MUP instructions may be largely eliminated from the software, leaving just the domain-specific knowledge. The increased complexity of the CPU should be more than offset by cutting out redundancy in the software, meaning an overall reduction in complexity---as suggested by the relatively small size in the figure of the SP machine, compared with the conventional computer.

\section{Potential benefits of the SP theory in some areas of application}\label{areas_of_application}

This section describes some areas of application for the SP theory, and its potential benefits, drawing on the `enabling technologies' summarised in Section \ref{sp_simplicity_power}. These examples are probably just the tip of the iceberg. To mix our metaphors, there is a rich vein to be mined.

Because of its importance, I believe it is worth repeating the principle, mentioned in Section \ref{philosophy_section}, that, even in specialised areas, a theory like the SP theory that has a broad base of support is likely to prove useful.

There seem to be two main reasons for this:

\begin{itemize}

\item A theory that scores well in terms of simplicity and power is likely to provide deeper insights than others. The germ theory of disease serves us better than the idea that diseases are caused by ``bad air''; recognising the role of oxygen in combustion is more practical than the phlogiston theory; and so on.

\item Even in one relatively narrow area of application, a broadly-supported theory is likely to promote the simplification and integration of structures and functions, both within the given area and between that area and others.

\end{itemize}

Even if existing technologies in a given area are doing well, there is likely to be a case for examining how the SP theory may be applied, and its potential benefits.

\subsection{Applications in the processing of natural language}\label{nl_applications}

As noted in Section \ref{nl_processing}, the SP system supports the parsing and production of language, it provides an elegant means of representing discontinuous dependencies in syntax, it is robust in the face of errors in data, it facilitates the integration of syntax with semantics, and it provides a role for context in the processing of language. There are many potential applications, including such things as checking for grammatical errors in text; predictive typing that takes account of grammar as well as the spellings of individual words; natural-language front-ends to databases; and the understanding and translation of natural languages.

It is true that some of these things are already done quite well with existing technologies but, as suggested earlier, there may nevertheless be advantages in exploring what can be done with SP system. It is likely, for example, to facilitate the integration of syntax with semantics, and it is likely to smooth the path for the integration of natural language processing with other aspects of intelligence: reasoning, learning, pattern recognition, and so on.

\subsubsection{Crowdsourcing as a route towards the understanding and translation of natural languages?}\label{crowdsourcing}

Given the apparent simplicity of representing syntactic and non-syntactic knowledge with patterns (Section \ref{representation_of_knowledge}), it may be feasible to use crowdsourcing in the manner of Wikipedia to create syntactic rules for at least one natural language and a realistically large set of ontologies to provide associated meanings.

With this kind of knowledge on the web and with a high-parallel, on-line version of the SP machine (as sketched in Section \ref{sp_machine}), it may be feasible to achieve understanding of natural language, at least at an elementary level.

With the same ontologies and with syntactic knowledge for two or more languages, it may be feasible to achieve translations between languages, with the ontologies functioning as an `interlingua'.\footnote{See, for example, `Interlingual machine translation', Wikipedia, \url{bit.ly/RlTcHU}.}

When translations are required amongst several different languages, this approach has the attraction that it is only necessary to create $n$ mappings, one between each of the $n$ languages and the interlingua. This contrasts with syntax-to-syntax systems such as Google Translate, where $n! / 2(n - 2)!$ pairs of languages are needed.

Another possible advantage is that, arguably, machine translation using an interlingua has the potential to produce translations of a better quality than may be achieved with syntax-to-syntax translation.

\subsection{Towards a versatile intelligence for autonomous robots}\label{autonomous_robots}

If a robot is to survive and be effective in a challenging environment like Mars, where communication lags restrict the help that can be provided by people, there are advantages if it can be provided with human-like intelligence and versatility---as much as possible. A theory like the SP theory, which provides a unified view of several different aspects of intelligence, is a good candidate for the development of that versatile intelligence.\footnote{This reference to autonomous robots does not in any way endorse or defend the irresponsible use of airborne `drones' or other autonomous robots for military or related kinds of operations, including the use of any such device without robust safeguards and controls.}

\subsection{Computer vision}\label{computer_vision}

Despite the problems noted in Section \ref{unfinished_business}, I believe the SP system has potential in the area of computer vision:

\begin{itemize}

\item It has potential to simplify and integrate several areas in computer vision, including feature detection and alignment, segmentation, deriving structure from motion, stitching of images together, stereo correspondence, scene analysis, and object recognition \citep[see, for example,][]{szeliski_2011}. With regard to the last two topics:

\begin{itemize}

\item Since scene analysis is, in some respects, similar to the parsing of natural language, and since the SP system performs well in parsing, it has potential in scene analysis as well. In the same way that a sentence may be parsed successfully without the need for explicit markers of the boundaries between successive words or between successive phrases, a scene may be analysed into its component parts without the need for explicit boundaries between objects or other elements in a scene.

\item The system provides a powerful framework for pattern recognition, as outlined in Section \ref{pattern_recognition}. It seems likely that this can be generalised for object recognition.

\end{itemize}

\item The SP system has potential for unsupervised learning of the knowledge required for recognition. For example, discrete objects may be identified by the matching and merging of patterns within stereo images ({\em cf.} \citet{marr_poggio_1979})\footnote{As with information compression, a focus on the isolation of discrete objects in binocular vision is distinct from the more usual interest in the way that slight differences between the two images enables us to see the scene in depth.} or within successive frames in a video, in much the same way that the word structure of natural language may be discovered via the matching and unification of patterns (Section \ref{unsupervised learning}). The system may also learn such things as classes of entity, and associations between entities, such as the association between black clouds and rain.

\item The system is likely to facilitate the seamless integration of vision with other aspects of intelligence: reasoning, planning, problem solving, natural language processing, and so on.

\item As noted earlier, the system is robust in the face of errors of omission, commission or substitution---an essential feature of any system that is to achieve human-like capabilities in vision.

\item There are potential solutions to the problems outlined in Section \ref{unfinished_business}:

\begin{itemize}

\item It is likely that the framework can be generalised to accommodate patterns in two dimensions.

\item As noted earlier, 3D structures may be modelled using 2D patterns, somewhat in the manner of architects' drawings. Knowledge of such structures may be built via the matching and unification of partially-overlapping 2D views.

\item The framework has potential to support both the discovery and recognition of low-level perceptual features.

\end{itemize}

\end{itemize}

\subsection{A versatile model for intelligent databases}\label{intelligent_databases}

As described in \citet{wolff_sp_intelligent_database}, the SP system provides a versatile framework for the creation of intelligent databases:

\begin{itemize}

\item It has the versatility to support the storage and retrieval of several kinds of knowledge, including object-oriented structures with inheritance of attributes (including multiple inheritance), relational tuples, and data that is structured as networks or trees.

\item As noted earlier, it provides for the integration of class-inclusion hierarchies with part-whole hierarchies.

\item It is free from the anomalies and inefficiencies that arise in existing systems that use a relational database as a means of creating object-oriented or entity-relationship data structures \citep[see][Section 4.2.3]{wolff_sp_intelligent_database}.

\item It provides for retrieval of information in the manner of query-by-example but has the potential to support the creation of query languages like SQL.

\item Unlike conventional databases, it supports various kinds of reasoning, and other kinds of `intelligence'.

\end{itemize}

\subsection{Maintaining multiple versions of a document or web page}

The multiple alignment concept provides an elegant means of maintaining multiple versions of any document or web page, such as, for example, versions in different languages. Parts which are shared amongst different versions---such as pictures or diagrams---may be kept separate from parts that vary amongst different versions---such as text in a specific language. As with other kinds of data, the framework provides the means of maintaining hierarchies of classes, with cross classification if that is required, and of maintaining part-whole hierarchies and their integration with class hierarchies.

Updates to high-level classes appear directly in all lower-level classes without the need for information to be copied amongst different versions.

\subsection{Software engineering: object-oriented design, verification and validation}

For the following main reasons, the SP system has potential in software engineering:

\begin{itemize}

\item Databases provide the mainstay of many software engineering projects and, as we have seen (Section \ref{intelligent_databases}), the SP system has potential in that area.

\item As noted in Section \ref{intelligent_databases}, the SP system provides for object-oriented design with inheritance of attributes, including multiple inheritance, and it provides `intelligence' as well.

\item It appears that the SP system may serve to model real-world procedures in business, administration, the law, and so on---although this has not yet been examined in any detail.

\end{itemize}

As was suggested in Section \ref{simplification_section}, the SP system has potential to reduce or eliminate `MUP' instructions and their repetition in different applications. To the extent that that proves possible, we may reduce the opportunities for errors to be introduced. There is potential to reduce or eliminate the need for `verification' in software development, with corresponding improvements in the quality of software.

The SP machine also has potential in `validation': helping to ensure that what is being developed is what is required. If domain-specific knowledge, including the requirements for a software system, can be loaded directly into an SP machine, without the need for the traditional kind of programming, this will help to ensure that what the system does is what the users want. There is potential for users to have tighter control than is traditional over what a system does, with more freedom to make changes when required.\footnote{Notice that the distinction, just referenced, between domain-specific knowledge and MUP instructions is different from the often-made distinction between declarative programming and imperative or procedural programming. It is envisaged that the SP machine would reduce or eliminate the MUP instructions but that it would be able to model both the declarative `what' and the procedural `how' of the real world.}

\subsection{Management of data and the drawing of inferences in criminal investigations}

It seems likely that an industrial-strength version of the SP machine would be useful in crime investigations:

\begin{itemize}

\item It would provide a means of storing and managing the data that are gathered in such investigations, often in large amounts.

\item It has potential to recognise similarities between a given crime and other crimes, either current or past---and likewise for suspects.

\item It may prove useful in piecing together coherent patterns from partially-overlapping fragments of information, in much the same way that partially-overlapping digital photographs may be stitched together to create a larger picture.

\item Given the potential of the SP system for different kinds of reasoning, including reasoning with information that is not complete, the system may prove useful in suggesting avenues to be explored, perhaps highlighting aspects of an investigation that might otherwise be overlooked.

\end{itemize}

\subsection{Managing `big data' and gaining value from it}\label{big_data}

Because information compression is central to the workings of the SP system, and because the system has potential to deliver relatively high levels of compression (Section \ref{compression_potential}), it may prove useful in reducing the sizes of the very large data sets known as `big data', and thus helping to make them more manageable.

The capabilities of the SP system in pattern recognition are potentially useful in the process of searching big data for particular kinds of information.

With solutions to the residual problems in unsupervised learning (Section \ref{unfinished_business}, it seems likely that the SP machine, via the `DONSVIC' principle (Section \ref{unsupervised learning}), will prove useful in discovering `interesting' or `useful' structures in big data, including significant entities and classes of entity, and rules or regularities, including discontinuous dependencies in data.

Since the concept of multiple alignment in the SP theory is a modified version of that concept in bioinformatics, the SP machine may prove useful in the discovery or detection of interesting structures in DNA sequences or amino-acid sequences.

\subsection{Knowledge, reasoning, and the semantic web}\label{semantic_web}

The SP framework may make a useful contribution to the development of the `semantic web'---a `web of data' to provide machine-understandable meaning for web pages \citep[see, for example,][]{bernerslee_etal_2001,shadbolt_hall_bernerslee_2006}.\footnote{See also \url{www.w3.org/standards/semanticweb/}.} In this connection, the main attractions of the system appear to be:

\begin{itemize}

\item {\em Simplicity, versatility and integration in the representation of knowledge}. As noted in Section \ref{representation_of_knowledge}, the SP system combines simplicity in the underlying format for knowledge with the versatility to represent several different kinds of knowledge---and it facilitates the seamless integration of different kinds of knowledge.

\item {\em Versatility in reasoning}. The system provides for several kinds of reasoning and their inter-working (Section \ref{reasoning_section}).

\item {\em Potential for natural language understanding}. As noted in Section \ref{nl_processing}, seamless integration of syntactic and semantic knowledge in the SP system gives it potential for the understanding of natural language. If that proves possible, semantic structures may be derived from textual information in web pages, without the need, in those web pages, for separate provision of ontologies or the like.

\item {\em Potential for automatic learning of ontologies}. As was suggested in Section \ref{big_data}, the SP system has potential, via the `DONSVIC' principle, for the extraction of interesting or useful structures from big data. These may include the kinds of ontologies which have been a focus of interest in work on the development of the semantic web.

\item {\em Uncertainty and vagueness}. The SP system is inherently probabilistic and it is robust in the face of incomplete information and errors of commission or substitution. These capabilities appear promising as a means of coping with uncertainty and vagueness in the semantic web \citep{lukasiewicz_straccia_2008}.

\end{itemize}

\subsection{Medical diagnosis}\label{medical_diagnosis}

The way in which the SP system may be applied in medical diagnosis is described in \citet{wolff_medical_diagnosis}. The potential benefits of the SP system in that area of application include:

\begin{itemize}

\item A format for representing diseases that is simple and intuitive.

\item An ability to cope with errors and uncertainties in diagnostic information.

\item The simplicity of storing statistical information as frequencies of occurrence of diseases.

\item The system provides a method for evaluating alternative diagnostic hypotheses that yields true probabilities.

\item It is a framework that should facilitate the unsupervised learning of medical knowledge and the integration of medical diagnosis with other AI applications.

\end{itemize}

The main emphasis in \citet{wolff_medical_diagnosis} is on medical diagnosis as pattern recognition. But the SP system has potential for causal reasoning \citep[see][Section 7.9]{wolff_2006} which, in a medical context, may enable the system to make probabilistic inferences such as ``The patient's fatigue may be caused by anaemia which may be caused by a shortage of iron in the diet''.

\subsection{Detection of computer viruses}\label{computer_viruses}

The detection of already-known computer viruses and other malware can be more subtle and difficult than simply looking for exact matches for the `signatures' of viruses. The offending code may be rather similar to perfectly legitimate code and it may be contained within a compressed (`packed') executable file.

Here, the SP system and its capabilities for pattern recognition and learning may prove useful:

\begin{itemize}

\item Recognition in the SP system is probabilistic, it does not depend on the presence or absence of any particular feature or combination of features, and it can cope with errors.

\item By compressing known viruses into a set of SP patterns, the system can reduce the amount of information needed to specify viruses and, as a consequence, it can reduce the amount of processing needed for the detection of viruses.

\end{itemize}

\subsection{Support for the economical transmission of data}

A compressed version of a body of information, $I$, may be seen to comprise two parts:

\begin{itemize}

\item A `grammar' for $I$ containing patterns, at one or more levels of abstraction, that occur repeatedly in $I$.

\item An `encoding' of $I$ in terms of the grammar, including non-repeating information in $I$.

\end{itemize}

Where there is significant redundancy in $I$---which is true of most kinds of natural-language text and most kinds of images---the encoding is likely to be much smaller than the grammar. A grammar for $I$ may also support the economical encoding of other bodies of information, provided they contain the same kinds of structures as $I$.

These things can provide the means of transmitting information very economically:

\begin{itemize}

\item A receiver (such as a TV set) may be equipped with a grammar for the kind of information it is designed to receive (eg, TV images), and some version of the SP system so that it may decode incoming information in terms of the grammar.

\item Instead of transmitting `raw' data, or even data that has been compressed in the usual manner (containing both the grammar and the encoding), the encoding by itself would be sufficient.

\item The original data may be reconstructed fully, without any loss of information, by decoding the transmitted information in terms of the stored grammar.

\end{itemize}

A simple version of this idea, using a dictionary of words as a `grammar' for ordinary text, is already recognised (see, for example, \citet{giltner_etal_1983}, \citet[Chapter 1]{storer_1988}). The potential advantage of the SP system is the creation of more sophisticated grammars yielding higher levels of compression, and the application of the idea to kinds of information---such as images---where an ordinary dictionary would not work.

Since ordinary TV works well without this kind of mechanism, some people may argue that there is no need for anything different. But:

\begin{itemize}

\item The growing popularity of video, TV and films on mobile services is putting pressure on mobile bandwidth.\footnote{See, for example, ``Data jam threat to UK mobile networks'', BBC News, 2012-11-16, \href{http://bbc.in/T5g14s}{bbc.in/T5g14s}.}

\item It is likely that some of the bandwidth for terrestrial TV will be transferred to mobile services,\footnote{See, for example, ``TVs will need retuning again to make room for mobile services'', The Guardian, 2012-11-16, \href{http://bit.ly/ZZ9N8C}{bit.ly/ZZ9N8C}.} creating an incentive to use the remaining bandwidth efficiently.

\item There would be benefits in, for example, the transmission of information from a robot on Mars, or any other situation where a relatively large amount of information needs to be transmitted as quickly as possible over a relatively low-bandwidth channel.

\end{itemize}

\subsection{Data fusion}

There are many situations where it can be useful to merge or integrate two or more sources of information, normally when there is some commonality amongst the different sources. For example, in ecological research there may be a need to integrate different sources of information about the movement of whales or other creatures around the world; and in the management of a website, there may be a need to consolidate a set of web pages that are saying similar things.

At first sight, the SP system appears to be tailor made for this area of application. After all, the merging of fully or partially matching patterns lies at the heart of the SP system. And for certain kinds of data fusion projects, that kind of capability could indeed prove to be effective and useful.

However, for better results, it is likely that solutions will be needed for the problems outlined in Section \ref{unfinished_business}. And, to rival human capabilities in the consolidation of natural language texts, the potential of the system for the understanding of natural language would need to be developed to the point where it could understand that two bodies of text are saying the same thing but with different words.

\section{Conclusion}

The potential benefits of the SP theory and SP machine include simplification and integration of concepts in computer science (including artificial intelligence), and simplification of computing systems, including software.

As a theory with a broad base of support, the SP theory is likely to provide useful insights in many areas and to promote the integration of structures and functions, both within a given area and amongst different areas.

The SP theory promises new and improved solutions to problems in several areas including natural language processing (with potential for the understanding and translation of natural languages), the need for a versatile intelligence in autonomous robots, computer vision, intelligent databases, maintaining multiple versions of documents or web pages, software engineering, criminal investigations, the management of big data and gaining benefits from it, the semantic web, medical diagnosis, the detection of computer viruses, the economical transmission of data, and data fusion.

A useful step forward in the development of these ideas would be the creation, as a software virtual machine, of a high-parallel, web-based, open-source version of the SP machine, taking advantage of powerful search mechanisms in one of the existing search engines, and with a good user interface. This would provide a means for researchers to explore what can be done with the system and to refine it.


\begin{thebibliography}{34}
\providecommand{\natexlab}[1]{#1}
\providecommand{\url}[1]{\texttt{#1}}
\expandafter\ifx\csname urlstyle\endcsname\relax
  \providecommand{\doi}[1]{doi: #1}\else
  \providecommand{\doi}{doi: \begingroup \urlstyle{rm}\Url}\fi

\bibitem[Anderson et~al.(2004)Anderson, Bothell, Byrne, Douglass, Lebiere, and
  Qin]{anderson_etal_2004}
J.~R. Anderson, D.~Bothell, M.~D. Byrne, S.~Douglass, C.~Lebiere, and Y.~Qin.
\newblock An integrated theory of the mind.
\newblock \emph{Psychological Review}, 111\penalty0 (4):\penalty0 1036--1060,
  2004.

\bibitem[Attneave(1954)]{attneave_1954}
F.~Attneave.
\newblock Some informational aspects of visual perception.
\newblock \emph{Psychological Review}, 61:\penalty0 183--193, 1954.

\bibitem[Barlow(1959)]{barlow_1959}
H.~B. Barlow.
\newblock Sensory mechanisms, the reduction of redundancy, and intelligence.
\newblock In {HMSO}, editor, \emph{The Mechanisation of Thought Processes},
  pages 535--559. Her Majesty's Stationery Office, London, 1959.

\bibitem[Barlow(1969)]{barlow_1969}
H.~B. Barlow.
\newblock Trigger features, adaptation and economy of impulses.
\newblock In K.~N. Leibovic, editor, \emph{Information Processes in the Nervous
  System}, pages 209--230. Springer, New York, 1969.

\bibitem[Barrow(1992)]{barrow_1992}
J.~D. Barrow.
\newblock \emph{Pi in the Sky}.
\newblock Penguin Books, Harmondsworth, 1992.

\bibitem[Berners-Lee et~al.(2001)Berners-Lee, Hendler, and
  Lassila]{bernerslee_etal_2001}
T.~Berners-Lee, J.~Hendler, and O.~Lassila.
\newblock The {S}emantic {W}eb.
\newblock \emph{Scientific American}, pages 35--43, May 2001.

\bibitem[Giltner et~al.(1983)Giltner, Mueller, and Fiest]{giltner_etal_1983}
M.E. Giltner, J.~C. Mueller, and R.~R. Fiest.
\newblock Data compression, encryption, and in-line transmission system.
\newblock United States Patent number 4386416, 1983.
\newblock Filed: 2 June 1980; Issued: 31 May 1983.

\bibitem[Gold(1967)]{gold_1967}
M.~Gold.
\newblock Language identification in the limit.
\newblock \emph{Information and Control}, 10:\penalty0 447--474, 1967.

\bibitem[Iwanska and Zadrozny(1997)]{iwanska_zadrozny_1997}
L.~Iwanska and W.~Zadrozny.
\newblock Introduction to special issue on context in natural language
  processing.
\newblock \emph{Computational Intelligence}, 13\penalty0 (3):\penalty0
  301--308, 1997.

\bibitem[Laird(2012)]{laird_2012}
J.~E. Laird.
\newblock \emph{The Soar Cognitive Architecture}.
\newblock The MIT Press, Cambridge, Mass., 2012.
\newblock ISBN-13: 978-0-262-12296-2.

\bibitem[Li and Vit\'{a}nyi(2009)]{li_vitanyi_2009}
M.~Li and P.~Vit\'{a}nyi.
\newblock \emph{An Introduction to Kolmogorov Complexity and Its Applications}.
\newblock Springer, New York, 2009.

\bibitem[Lukasiewicz and Straccia(2008)]{lukasiewicz_straccia_2008}
T.~Lukasiewicz and U.~Straccia.
\newblock Managing uncertainty and vagueness in description logics for the
  {S}emantic {W}eb.
\newblock \emph{Proceedings of the Royal Society, Series B}, 6\penalty0
  (4):\penalty0 291--308, 2008.

\bibitem[Marr and Poggio(1979)]{marr_poggio_1979}
D.~Marr and T.~Poggio.
\newblock A computational theory of human stereo vision.
\newblock \emph{Proceedings of the Royal Society of London. Series B},
  204\penalty0 (1156):\penalty0 301--328, 1979.

\bibitem[Newell(1973)]{newell_1973}
A.~Newell.
\newblock You can't play 20 questions with nature and win: projective comments
  on the papers in this symposium.
\newblock In W.~G. Chase, editor, \emph{Visual Information Processing}, pages
  283--308. Academic Press, New York, 1973.

\bibitem[Newell(1990)]{newell_1990}
A.~Newell, editor.
\newblock \emph{Unified Theories of Cognition}.
\newblock Harvard University Press, Cambridge, Mass., 1990.

\bibitem[Newell(1992)]{newell_1992}
A.~Newell.
\newblock Pr\'ecis of {\em \uppercase{u}nified \uppercase{t}heories of
  \uppercase{c}ognition}.
\newblock \emph{Behavioural and Brain Sciences}, 15\penalty0 (3):\penalty0
  425--437, 1992.

\bibitem[Oliva and Torralba(2007)]{oliva_torralba_2007}
A.~Oliva and A.~Torralba.
\newblock The role of context in object recognition.
\newblock \emph{Trends in cognitive sciences}, 11\penalty0 (12):\penalty0
  520--527, 2007.

\bibitem[Post(1943)]{post_1943}
E.~L. Post.
\newblock Formal reductions of the general combinatorial decision problem.
\newblock \emph{American Journal of Mathematics}, 65:\penalty0 197--268, 1943.

\bibitem[Prince(2012)]{prince_2012}
S.~J.~D. Prince.
\newblock \emph{Computer Vision: Models, Learning, and Inference}.
\newblock Cambridge University Press, Cambridge, 2012.
\newblock ISBN 978-1-107-01179-3.

\bibitem[Rissanen(1978)]{rissanen_1978}
J.~Rissanen.
\newblock Modelling by the shortest data description.
\newblock \emph{Automatica-J, {IFAC}}, 14:\penalty0 465--471, 1978.

\bibitem[Schmidhuber et~al.(2011)Schmidhuber, Th\'{o}risson, and
  Looks]{agi_2011}
J.~Schmidhuber, K.~R. Th\'{o}risson, and M.~Looks, editors.
\newblock \emph{Proceedings of the Fouth International Conference on Artificial
  General Intelligence (AGI 2011)}, volume 6830 of \emph{Lecture Notes in
  Artificial Intelligence}, 2011. Springer.
\newblock ISBN 978-3-642-22886-5.

\bibitem[Shadbolt et~al.(2006)Shadbolt, Hall, and
  Berners-Lee]{shadbolt_hall_bernerslee_2006}
N.~Shadbolt, W.~Hall, and T.~Berners-Lee.
\newblock The semantic web revisited.
\newblock \emph{{IEEE} Intelligent Systems}, 21\penalty0 (3):\penalty0 96--101,
  2006.

\bibitem[Solomonoff(1964)]{solomonoff_1964}
R.~J. Solomonoff.
\newblock A formal theory of inductive inference. {P}arts {I} and {II}.
\newblock \emph{Information and Control}, 7:\penalty0 1--22 and 224--254, 1964.

\bibitem[Storer(1988)]{storer_1988}
J.~A. Storer.
\newblock \emph{Data Compression: Methods and Theory}.
\newblock Computer Science Press, Rockville, Maryland, 1988.

\bibitem[Szeliski(2011)]{szeliski_2011}
R.~Szeliski.
\newblock \emph{Computer Vision: Algorithms and Applications}.
\newblock Springer, London, 2011.
\newblock ISBN 978-1-84882-934-3.

\bibitem[Turing(1936)]{turing_1936}
A.~M. Turing.
\newblock On computable numbers with an application to the
  {E}ntscheidungsproblem.
\newblock \emph{Proceedings of the London Mathematical Society}, 42:\penalty0
  230--265 and 544--546, 1936.

\bibitem[Turing(1950)]{turing_1950}
A.~M. Turing.
\newblock Computing machinery and intelligence.
\newblock \emph{Mind}, 59:\penalty0 433--460, 1950.

\bibitem[Wallace and Boulton(1968)]{wallace_boulton_1968}
C.~S. Wallace and D.~M. Boulton.
\newblock An information measure for classification.
\newblock \emph{Computer Journal}, 11\penalty0 (2):\penalty0 185--195, 1968.

\bibitem[Wolff(1977)]{wolff_1977}
J.~G. Wolff.
\newblock The discovery of segments in natural language.
\newblock \emph{British Journal of Psychology}, 68:\penalty0 97--106, 1977.
\newblock See: \url{www.cognitionresearch.org/lang\_learn.html\#wolff\_1977}.

\bibitem[Wolff(1988)]{wolff_1988}
J.~G. Wolff.
\newblock Learning syntax and meanings through optimization and distributional
  analysis.
\newblock In Y.~Levy, I.~M. Schlesinger, and M.~D.~S. Braine, editors,
  \emph{Categories and Processes in Language Acquisition}, pages 179--215.
  Lawrence Erlbaum, Hillsdale, NJ, 1988.
\newblock See: \url{www.cognitionresearch.org/lang\_learn.html\#wolff\_1988}.

\bibitem[Wolff(1990)]{wolff_1990}
J.~G. Wolff.
\newblock Simplicity and power---some unifying ideas in computing.
\newblock \emph{Computer Journal}, 33\penalty0 (6):\penalty0 518--534, 1990.
\newblock See: \url{www.cognitionresearch.org/papers/early/early.htm}.

\bibitem[Wolff(2006{\natexlab{a}})]{wolff_2006}
J.~G. Wolff.
\newblock \emph{Unifying Computing and Cognition: the {SP} Theory and Its
  Applications}.
\newblock CognitionResearch.org, Menai Bridge, 2006{\natexlab{a}}.
\newblock ISBNs: 0-9550726-0-3 (ebook edition), 0-9550726-1-1 (print edition).
  Distributors, including Amazon.com, are detailed on
  \url{www.cognitionresearch.org/books/sp_book/retailers.htm}. The publisher
  and its website was previously CognitionResearch.org.uk.

\bibitem[Wolff(2006{\natexlab{b}})]{wolff_medical_diagnosis}
J.~G. Wolff.
\newblock Medical diagnosis as pattern recognition in a framework of
  information compression by multiple alignment, unification and search.
\newblock \emph{Decision Support Systems}, 42:\penalty0 608--625,
  2006{\natexlab{b}}.
\newblock See:
  \url{www.cognitionresearch.org/papers/applications/medical/medical\_applications.htm}.

\bibitem[Wolff(2007)]{wolff_sp_intelligent_database}
J.~G. Wolff.
\newblock Towards an intelligent database system founded on the {SP} theory of
  computing and cognition.
\newblock \emph{Data \& Knowledge Engineering}, 60:\penalty0 596--624, 2007.
\newblock See: \url{www.cognitionresearch.org/papers/dbir/dbir.htm}.

\end{thebibliography}

\end{document}